\pdfoutput=1
\documentclass[11pt]{article}

\usepackage[preprint]{acl}

\usepackage{times}
\usepackage{latexsym}
\usepackage{hyperref}
\usepackage{color, colortbl}
\definecolor{Gray}{gray}{0.9}
\usepackage{array}
\usepackage{booktabs}
\usepackage{multirow}
\usepackage{amsmath}
\usepackage{amssymb}
\usepackage{graphicx}
\usepackage{enumerate}
\usepackage{diagbox}
\usepackage{mathtools}
\usepackage{paralist}
\usepackage{tabularx}
\usepackage{lipsum}
\usepackage{ragged2e}
\usepackage{makecell}

\usepackage[T1]{fontenc}
\usepackage[utf8]{inputenc}
\usepackage{csquotes}

\usepackage{microtype}

\usepackage{inconsolata}

\usepackage{graphicx}

\newcommand{\dataset}{\textsc{EcomScriptBench}}
\newcommand{\task}{\textsc{EcomScript}}

\newcommand{\emojiname}{\raisebox{-2.5pt}{\includegraphics[width=1.1em]{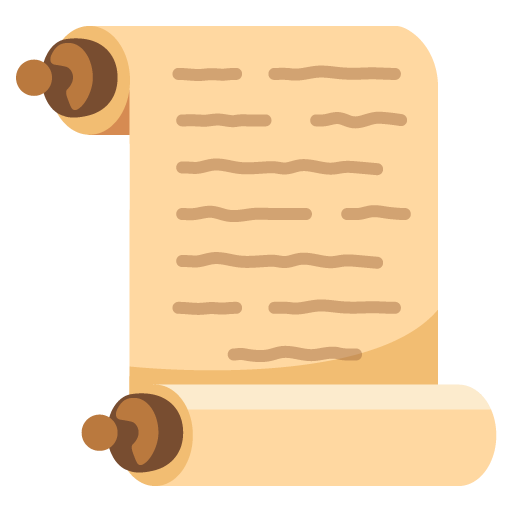}}\textsc{EcomScriptBench}}

\definecolor{colorxmark}{RGB}{255, 87, 51}
\definecolor{colorcmark}{RGB}{66, 154, 137}
\definecolor{headcolor}{HTML}{018161}
\definecolor{relationcolor}{HTML}{d95f02}
\definecolor{tailcolor}{HTML}{6560a3}

\title{\emojiname: A Multi-task Benchmark for E-commerce Script Planning via Step-wise Intention-Driven Product Association}

\author{Weiqi Wang\thanks{Work done during his internship at Amazon.com Inc.}$^{\spadesuit\clubsuit}$, 
Limeng Cui$^{\clubsuit}$,
Xin Liu$^{\clubsuit}$,
Sreyashi Nag$^{\clubsuit}$,
Wenju Xu$^{\clubsuit}$,
Chen Luo$^{\clubsuit}$,\\
\textbf{
Sheikh Sarwar$^{\clubsuit}$,
Yang Li$^{\clubsuit}$,
Hansu Gu$^{\clubsuit}$,
Hui Liu$^{\clubsuit}$,
Changlong Yu$^{\clubsuit}$,
Jiaxin Bai$^{\spadesuit}$,}\\
\textbf{
Yifan Gao$^{\clubsuit}$,
Haiyang Zhang$^{\clubsuit}$,
Qi He$^{\clubsuit}$,
Shuiwang Ji$^{\dagger\heartsuit\clubsuit}$,
Yangqiu Song\thanks{Visiting academic scholar at Amazon.com Inc.}$^{\spadesuit\clubsuit}$}\\
$^{\spadesuit}$Department of Computer Science and Engineering, HKUST, Hong Kong SAR, China\\
$^{\clubsuit}$Amazon.com Inc, Palo Alto, CA, USA\\
$^{\heartsuit}$Department of Computer Science \& Engineering, Texas A\&M University, Texas, USA\\
\texttt{\{wwangbw, yqsong\}@cse.ust.hk, \{culimeng, xliucr, cheluo\}@amazon.com}
}

\begin{document}
\maketitle
\begin{abstract}
Goal-oriented script planning, or the ability to plan coherent sequences of actions toward specific goals, is commonly used by humans to plan for daily activities. 
In e-commerce, customers increasingly seek LLM-based assistants to plan for them with a script and recommend products at each step, thereby facilitating convenient and efficient shopping experiences. 
However, this capability remains underexplored due to several challenges, including the inability of LLMs to simultaneously conduct script planning and product retrieval, difficulties in matching products caused by semantic discrepancies between planned actions and search queries, and a lack of methods and benchmark data for evaluation.
In this paper, we step forward by formally defining the task of E-commerce Script Planning (\task) as three sequential subtasks. 
We propose a novel framework that enables the scalable generation of product-enriched scripts by associating products with each step based on the semantic similarity between the actions and their purchase intentions. 
By applying our framework to real-world e-commerce data, we construct the very first large-scale~\task{} dataset, \dataset, which includes 605,229 scripts sourced from 2.4 million products. 
Human annotations are then conducted to provide gold labels for a sampled subset, forming an evaluation benchmark. 
Extensive experiments reveal that current (L)LMs face significant challenges with \task{} tasks, even after fine-tuning, while injecting product purchase intentions improves their performance.
\end{abstract}

\section{Introduction}

\begin{figure}[t]
     \centering
     \includegraphics[width=1\linewidth]{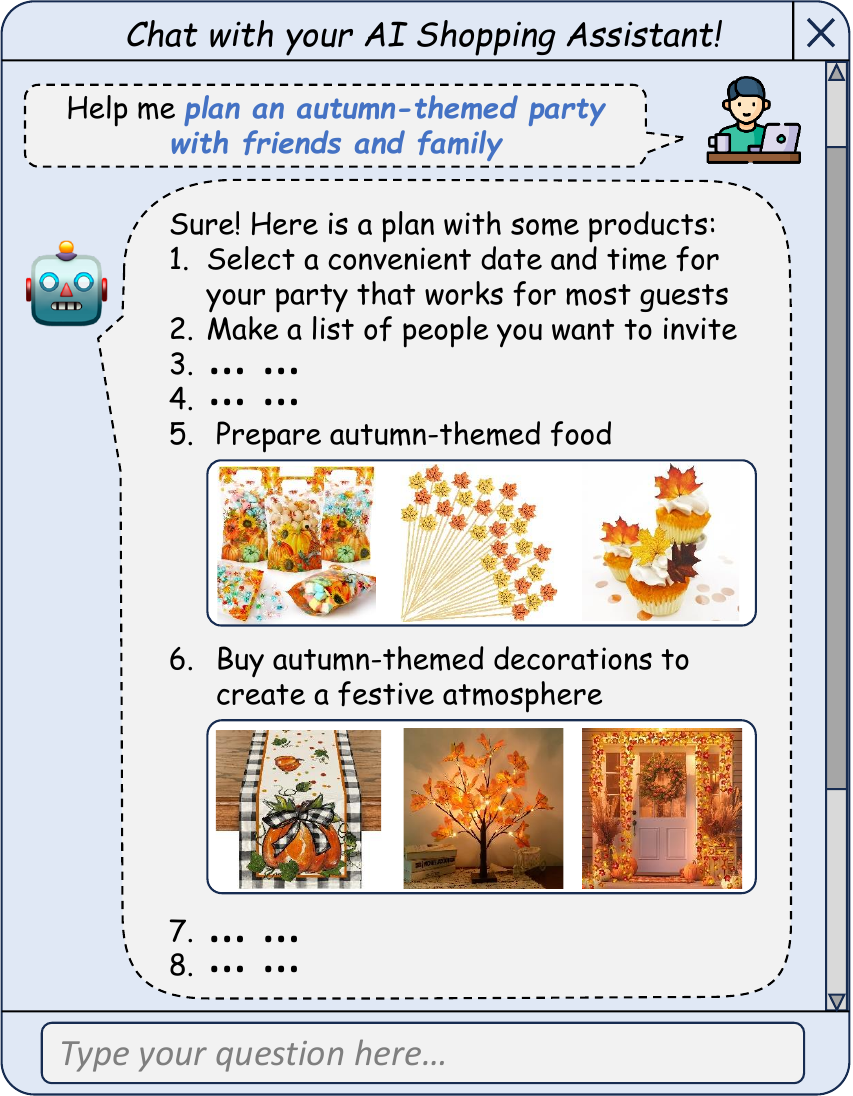}
     % \vspace{-0.2in}
     \caption{An example of a \textbf{\textit{product-enriched script}} for planning the objective of \textit{plan an autumn-themed party with friends and family}, with relevant products associated with some steps. 
     Note that for simpler steps, such as the first two, no products are needed.}
    \label{fig:introduction_example}
    % \vspace{-0.25in}
\end{figure}

In our daily lives, humans commonly plan a sequence of general prototypical actions, usually in the form of step-by-step instructions, to achieve a specific objective~\cite{abbott1985representation}. 
This capability, also known as \textbf{\textit{goal-oriented script planning}}~\cite{bower1979scripts,schank1975scripts}, forms the foundation of situationally grounded planning for complex scenarios, which is crucial for intelligent agents. 
With recent advances in Large Language Models (LLMs;~\citealp{GPT4o,GPT4omini,LLAMA3}), recent works~\cite{DBLP:conf/acl/YuanCFGSJXY23,DBLP:conf/acl/0005LCHH0J23,DBLP:conf/eacl/ChanCWJFLS24,DBLP:conf/emnlp/DengC00FZYS24} have demonstrated the strong capabilities of LLMs in script planning.
This has led to the development of LLM-based script planners across downstream domains~\cite{DBLP:conf/emnlp/HaoGMHWWH23,DBLP:conf/aaai/HazraMR24}.

In e-commerce, there is a growing trend of customers querying LLM-empowered shopping assistants to create scripts tailored to their specific needs or objectives, with each step featuring relevant products. 
For example, as illustrated in Figure~\ref{fig:introduction_example}, LLM assistant is expected to generate an eight-step script toward the user’s objective, \textit{plan an autumn-themed party with friends and family}, while also associating products that customers may find useful for achieving each step, such as items related to food and decoration in this example.
However, for simple actions that can be completed without additional items, no products are recommended. 
A goal-oriented script with product recommendations at certain steps that may necessitate product purchases, which we term as \textbf{product-enriched script}, facilitates a convenient shopping experience, saves customers from multiple rounds of searches, and ultimately fosters new concierge-style applications for LLM shopping assistants to promote business.

Despite its significant potential, the exploration of this ability has faced several challenges. 
First, while current LLMs excel in script planning, they struggle to retrieve relevant products from the vast pool in e-commerce platforms.
Although some LLMs have been pre-trained with e-commerce product knowledge, prior studies~\cite{DBLP:conf/aaai/LiMWHJ0X0J24,DBLP:conf/icml/0009LC0N24} indicate that they still face difficulties in generating precise product titles that accurately align with specific items in the pool for further retrieval.
Additionally, while a generate-then-retrieve approach--where LLMs first plan the script and then use the steps as corresponding search queries for product searches using traditional search engines--might be feasible, there is often a semantic gap between the planned steps (the actions users should take sequentially) and the search queries intended for search engines (see Section~\ref{appendix:semantic_gap_retrieval}). 
This gap arises because queries typically include product features and descriptions provided by users, which are matched against product metadata. 
When users search for actions that a product can facilitate, this discrepancy compromises the matching mechanism, further undermining the effectiveness of search engines and limiting their ability to address the shortcomings of LLMs in both planning and product retrieval.
Finally, there is a notable lack of methods and benchmark datasets that incorporate both script plans and products at the step level, which are necessary to evaluate the current capabilities of LLMs in this area.

To bridge these gaps, in this paper, we formally define the process of \textbf{E-commerce Script Planning} (\task) as a three-step discriminative process consisting of three sequential sub-tasks. 
We then introduce a novel framework for automatically guiding LLMs in generating product-enriched scripts by incorporating product keyword filtering at each step to narrow the search scope for relevant products. 
To address semantic discrepancies, we search for products based on their \textit{purchase intentions}, which represent a customer's underlying motivation to buy the product~\cite{chang1994price}, and then filter out those whose intentions don't align closely with the action at each step.

By applying our framework to Amazon Review data~\cite{AmazonReview}, we construct a large-scale knowledge base,~\dataset, that includes 605,229 product-enriched scripts derived from real user purchase reviews, alongside 2.4 million products, each linked to ten distinct purchase intentions. 
Within each script, we associate up to three products with each step by applying our intention alignment strategy (\S~\ref{sec:step_intention_alignment}).
Human annotations are then conducted to provide gold labels for 15,000 randomly sampled entries across three subtasks, thereby constructing an evaluation benchmark. 
We then experiment with over 20 (L)LMs, applying both fine-tuning and advanced prompting techniques to \task{} tasks. 
Our findings reveal that all LMs encounter significant challenges in addressing these tasks. 
Further analysis identifies potential reasons for their underperformance and demonstrates that injecting purchase intentional knowledge significantly enhances LLMs' performance. 
% We will make our data and models publicly available upon acceptance.

\section{Related Works}
\subsection{Goal-oriented Script Planning}
Goal-oriented scripts refer to a coherent and appropriate sequence of steps, usually in the form of actions, as instructions for achieving a goal~\cite{DBLP:conf/acl/RegneriKP10}. 
They are a common reflection of language planning capabilities, often observed in embodied AI~\cite{DBLP:conf/icra/GanZSABGYDM0T22} and robotics~\cite{DBLP:conf/icra/ZhangTSMQSSZG24}.
In the era of LLMs, various works have explored their script planning capabilities.
\citet{DBLP:conf/acl/YuanCFGSJXY23} proposed an over-generate-then-filter framework to improve the constraint language planning capabilities of LLMs and distilled a knowledge base from it. 
\citet{DBLP:conf/eacl/SunXZJ23,DBLP:conf/acl/0005LCHH0J23} attempted generative script learning in a multimodal manner to enhance the planning abilities of large vision-language models. 
\citet{DBLP:conf/ijcai/JoshiAPM23} designed an interactive text-based gaming framework that consists of daily real-world human activities as another benchmark.
Nevertheless, none of the prior works have explored script planning in the context of e-commerce, which holds significant potential for customers wishing to plan toward their desired objectives and purchase necessary products at every step all at once.

\subsection{Purchase Intention Understanding}
Purchase intention represents the implicit mental state that motivates customers’ purchase behaviors~\cite{anscombe2000intention}, which simulates the underlying reasons a customer wishes to achieve with the purchase of a product~\cite{DBLP:conf/emnlp/ChanJYDF0L0WS24}.
Various existing works have already examined the impact of consumer shopping intentions on downstream applications~\cite{DBLP:conf/www/DaiZNWWL06,DBLP:conf/www/ZhangFDY16,DBLP:conf/wsdm/HaoHPWYWW22,MIKO}.
Specifically,~\citet{DBLP:conf/emnlp/NiLM19} collected real-world customer reviews to investigate the underlying purchase intentions in consumer purchase behavior and created a large-scale review dataset based on Amazon.
\citet{FolkScope,COSMO,DBLP:journals/corr/abs-2412-11500} then leveraged this data and proposed a semi-supervised intention generation framework to obtain purchase intentions at scale (FolkScope and COSMO) by distilling OPT~\cite{OPT}.
\citet{MIND} further strengthened this approach by incorporating visual signals from product images to guide the generation of more feature-oriented purchase intentions that align with stronger human preferences.
\citet{IntentionQA} transformed FolkScope into an evaluation benchmark and demonstrated that LLMs cannot effectively utilize intention for product recommendation.
In our work, we share a similar aspiration of using purchase intention as the key to match products that best align with each actionable step in every script, enabling LLMs to implicitly leverage intention for product retrieval.

\section{Problem Definition}
\subsection{\task{} Task Definitions}
\label{sec:task_definitions}
We first introduce our definition of the proposed e-commerce script planning tasks (\task). 
Since both asking an LLM to generate a script with products associated with each step and evaluating such generations are difficult to accomplish directly, it is challenging to formulate the task simply as a one-step generative task and evaluate it in an open-ended manner. 
To this end, we propose three sequential discriminative tasks to emulate the process, with the aspiration that an LLM can perform these three tasks to build a generate-then-discriminate paradigm that fully enables automated e-commerce script planning. 
Initially, the model is given a user objective $o$, a script consisting of $k$ steps toward this objective $S_o=\{s_1,s_2,...,s_k\}$ (collected from the user or generated by the LLM), and a pool of $n$ e-commerce products $P=\{p_1,p_2,...,p_n\}$.

\noindent\textbf{Task 1: Script Verification}: The first task asks the model to determine the plausibility and feasibility of the script based on the given objective.
It gives the model $o$ and $S_o$ as input and requires the model to output a binary score $T_1(o, S_o)\in\{0, 1\}$ as the indicator where 1 indicates that the script is plausible and feasible, and 0 indicates otherwise.

\noindent\textbf{Task 2: Step-Product Discrimination}: The second task aims to determine whether a step in the script requires the purchase of a product to assist the user in accomplishing that step. 
If so, the model will then be provided with a product and asked to determine whether purchasing this product can help with the step. 
Formally, the model takes $o$, $s_i$, and $p_i$ as inputs and is required to output a binary score $T_2(o, s_i, p_i)\in\{0, 1\}$, where 0 indicates that the step does not require any product purchase or that the product cannot help, and 1 indicates that the product is a good match to contribute to the step.

\noindent\textbf{Task 3: Script-Products Verification}: The final task aims to determine the overall feasibility of a product-enriched script by providing the model with the objective, the script, and products associated with each step.
Formally, the model takes $o$, $S_o=\{s_1,s_2,...,s_k\}$, and the products at each step $P_{s_1}, P_{s_2}, ..., P_{s_k}$ as inputs and is expected to output a binary score $T_3(o, S_o, (P_{s_1}, P_{s_2}, ..., P_{s_k}))\in\{0, 1\}$ where 0 indicates that there are internal conflicts between different steps and products, while 1 means that all products are suitable for each step and can collaborate within the entire script.

The rationale behind this task design is that, with the filtering models associated with these three tasks and an LLM as the core shopping assistant, we can automate the process of e-commerce script planning. 
This is achieved by first asking the LLM to generate a script based on the user’s provided objective. 
Then, the \textbf{script verifier ($T_1$)} can determine the plausibility of the script and guide the LLM to improve it if necessary. 
Products will be retrieved according to our proposed step-intention alignment strategy, as explained later in Section~\ref{sec:step_intention_alignment}.
The \textbf{step-product discriminator ($T_2$)} can verify the results of each retrieved product associated with each step and remove unnecessary products for simple steps. 
Finally, the \textbf{script-products verifier ($T_3$)} will check the product-enriched script and ensure that all products can coordinate smoothly within the script to be recommended to the customer.

\begin{figure*}[t]
     \centering
     \includegraphics[width=\linewidth]{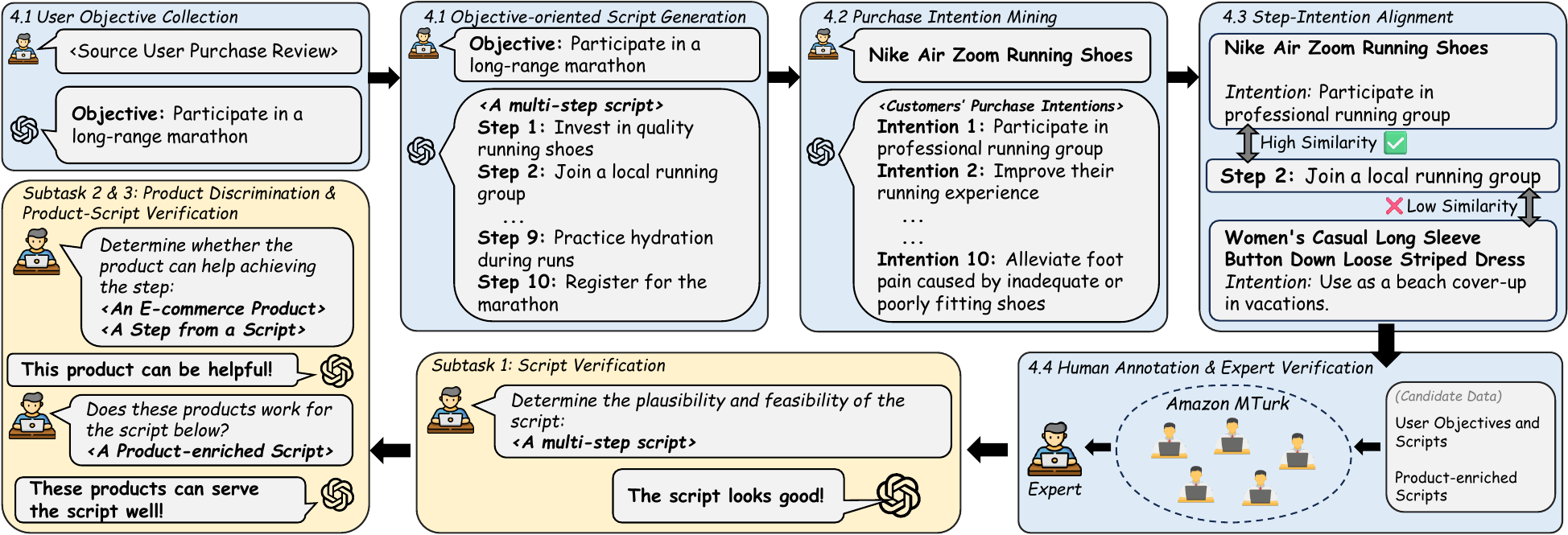}
     % \vspace{-0.3in}
     \caption{An overview of our benchmark curation and evaluation pipeline for~\dataset.}
    \label{fig:benchmark_curation_overview}
    % \vspace{-0.1in}
\end{figure*}

\subsection{Datasets}
\label{sec:datasets}
To ensure the practicality of applying our framework to real-world scenarios, we collect real world products from purchases made at Amazon.com.
% we use the Amazon Review Dataset~\cite{AmazonReview}, which consists of real-world product metadata (titles, features, descriptions, etc.) and user reviews collected after purchase from Amazon.com. 
To manage the overwhelming size of the product pool $P$ and reduce product redundancy, we randomly sample 10\% of unique products from each category while maintaining the original distribution. 
As a result, 2.4 million products and 3.7 million associated reviews are used for constructing~\dataset{}.

\subsection{Semantic Gaps in Product Retrieval}
\label{appendix:semantic_gap_retrieval}
Prior to this work, we conducted a preliminary pilot study to investigate the effectiveness of using search engines—considered a traditional alternative method—for retrieving products based on user-provided steps in the context of e-commerce script planning.
In this study, we selected 200 scripts at random and used their individual steps as search queries. 
The results showed that roughly 68\% of these queries returned only a limited assortment of products, indicating that search engines struggled to align product titles and metadata with the nuanced, natural language requirements expressed in the user queries.
We also observe that most of retrieved products are identical or very similar to each other, limiting the divergence of product association results.

For example, when searching for ``a reusable bottle that is easy to clean and suitable for carrying both hot and cold beverages,'' generic listings of reusable bottles were returned, with little emphasis on the specific attributes mentioned. 
This illustrates a semantic gap between how users describe their needs and the structured metadata currently used to index and retrieve products.
Addressing this gap will require incorporating richer contextual signals and better capturing user intent. 
By enhancing the information associated with products—such as their features, use cases, and suitability—systems can more effectively match user needs with relevant product recommendations.
This supports the main motivation of our study, which is to compensate for the weaknesses of semantic retrieval.

\section{\dataset{} Construction}
In this section, we introduce our method for synthesizing product-enriched scripts to construct an evaluation benchmark. 
An overview of the framework is shown in Figure~\ref{fig:benchmark_curation_overview}. 
Specifically, our framework consists of four main stages: (1) user objective and script collection, (2) product purchase intention mining, (3) script-product association through step-intention alignment, and (4) human annotation.

To enable scalable data collection, we use GPT-4o-mini~\cite{GPT4omini}, a powerful yet cost-efficient proprietary LLM, as the generator to collect user objectives, scripts, and product purchase intentions.
Following~\citet{FewShot} and~\citet{CANDLE}, we guide each generation stage with a few-shot prompt as described below:
\vspace{-0.1in}
\begin{center}
\resizebox{1\linewidth}{!}{
\begin{tabular}{l}
\textbf{\texttt{<TASK-PROMPT>}}\\
\textbf{\texttt{\textcolor{headcolor}{<INPUT$_1$>}\textcolor{tailcolor}{<OUTPUT$_{(1,1)}$>}}}~~\ldots~~\textbf{\texttt{\textcolor{tailcolor}{<OUTPUT$_{(1,N_1)}$>}}}\\
\textbf{\texttt{\textcolor{headcolor}{<INPUT$_2$>}\textcolor{tailcolor}{<OUTPUT$_{(2,1)}$>}}}~~\ldots~~\textbf{\texttt{\textcolor{tailcolor}{<OUTPUT$_{(2,N_2)}$>}}}\\
\ldots \\
\textbf{\texttt{\textcolor{headcolor}{<INPUT$_{5}$>}\textcolor{tailcolor}{<OUTPUT$_{(5,1)}$>}}}~~\ldots~~\textbf{\texttt{\textcolor{tailcolor}{<OUTPUT$_{(5,N_{5})}$>}}}\\
\textbf{\texttt{\textcolor{headcolor}{<INPUT$_{6}$>}}}
\end{tabular}
}
\end{center}
where we modify \textbf{\texttt{<TASK-PROMPT>}} at each stage to provide different instructions that inform the LLM of the generation objective and incorporate five \textbf{\texttt{\textcolor{headcolor}{<INPUT$_{i}$>}}} and \textbf{\texttt{\textcolor{tailcolor}{<OUTPUT$_{i}$>}}} pairs as few-shot exemplars for demonstration (prompts in Appendix~\ref{appendix:dataset_construction_prompt}).

\subsection{User Objective and Script Collection}
\label{sec:user_objective_collection}
We start by collecting user objectives by instructing the LLM to extract and infer them from user purchase reviews, as these are most practically aligned with real-world use cases. 
To achieve this, we let \textbf{\texttt{<TASK-PROMPT>}} clarify the goal to the LLM, which involves generating an objective that the customer is trying to achieve based on a series of purchases and their reviews. 
Note that we explicitly ask the LLM to avoid generating overly simplistic objectives and to aim for complex ones that require multiple steps to complete, in order to facilitate further script planning. 
We then populate \textbf{\texttt{\textcolor{headcolor}{<INPUT$_{i}$>}}} and \textbf{\texttt{\textcolor{tailcolor}{<OUTPUT$_{i}$>}}} with five pairs of user purchase reviews and a list of comma-separated objectives inferred from the reviews by experts. 
The LLM is then expected to infer a list of user objectives from the last given customer purchase review (\textbf{\texttt{\textcolor{headcolor}{<INPUT$_{6}$>}}}). 
If the LLM believes that no objective can be inferred, ``None'' will be generated instead.
To ensure high quality, we discard reviews that are too short or contain excessive punctuation or hashtags. 

With these objectives, we further instruct the LLM to generate goal-oriented scripts based on them. 
We similarly modify the prompt to achieve this by changing \textbf{\texttt{<TASK-PROMPT>}} to instruct the LLM to devise a coherent and sequential plan of steps, in the form of a script, with all steps being actions that are commonly seen in usual cases. 
Specifically, we require the LLM to avoid generating overly simple actions and to maximize the necessity of purchases in each step by generating actions that may require items to complete, whenever possible. 
We then populate \textbf{\texttt{\textcolor{headcolor}{<INPUT$_{i}$>}}} and \textbf{\texttt{\textcolor{tailcolor}{<OUTPUT$_{i}$>}}} with five pairs of user objectives and their corresponding actionable scripts, written by experts, as exemplars. 
The LLM is then expected to generate the script for the last given user objective (\textbf{\texttt{\textcolor{headcolor}{<INPUT$_{6}$>}}}). 
For simplicity, we ask the LLM to generate scripts that contain no more than 10 steps, and longer scripts will be truncated to a maximum of 10 steps. 
The mined objectives and generated scripts will be candidate data for the first subtask.

\subsection{Purchase Intention Mining}
\label{sec:purchase_intention_mining}
We then collect purchase intentions for e-commerce products, aiming to leverage these intentions as a key to bridge products and actionable steps. 
The rationale behind this approach is that intentions commonly reflect what customers wish to achieve with their purchases, which intuitively aligns with the semantics of actionable steps. 
This alignment helps overcome the semantic discrepancy found in traditional search queries, which typically focus on product features and metadata.

To collect purchase intentions, we follow~\citet{FolkScope} and utilize LLMs to distill intentional knowledge. 
Specifically, we modify \textbf{\texttt{<TASK-PROMPT>}} to instruct the LLM to infer purchase intentions by reasoning about the customer’s motivations and desires. 
We emphasize modeling the customer’s mental state, using phrases like ``PersonX wants to buy this because'' or ``PersonX believes buying this can'' to guide the generation. 
We then populate \textbf{\texttt{\textcolor{headcolor}{<INPUT$_{i}$>}}} and \textbf{\texttt{\textcolor{tailcolor}{<OUTPUT$_{i}$>}}} with five pairs of purchased product metadata and expert-drafted customer intentions as examples. 
The model is then asked to generate purchase intentions for the last given product (\textbf{\texttt{\textcolor{headcolor}{<INPUT$_{6}$>}}}). 
For each product, we collect 10 intentions, resulting in a total of 24 million intentions for 2.4 million products.

\subsection{Step-Intention Alignment}
\label{sec:step_intention_alignment}
In this stage, we first ask the LLM to determine whether a product purchase is necessary for each step, in order to filter out trivial actions that can be performed directly by the user without additional support from any product, such as ``invite friends'' and ``check the calendar.'' 
If the LLM believes that additional product purchases are necessary, we further ask it to generate a list of keywords to describe the product as thoroughly as possible. 
We will then filter products that contain any of these keywords to narrow down our search scope.

To achieve this, we modify the \textbf{\texttt{<TASK-PROMPT>}} to include the descriptions above and populate \textbf{\texttt{\textcolor{headcolor}{<INPUT$_{i}$>}}} and \textbf{\texttt{\textcolor{tailcolor}{<OUTPUT$_{i}$>}}} with five pairs of actionable steps in a script, along with their associated purchase necessity and relevant product keywords. 
The model will then infer the purchase necessity and relevant product keywords for the last provided step in the script (\textbf{\texttt{\textcolor{headcolor}{<INPUT$_{6}$>}}}).

For every step deemed necessary for product purchases and their filtered products, we use SentenceBERT~\cite{DBLP:conf/emnlp/ReimersG19} to calculate the average embedding similarity between each step and the purchase intentions of each product. 
For each step, we rank all filtered products to select the top three that best align with the actionable step in the script, using a lower-bound similarity threshold of $\tau=0.45$ to control for relevance, which is determined based on our observation of the similarity distribution.
We limit our selection to a maximum of three products to reduce overlap and maintain a manageable dataset size. 
Each step and its selected products form the candidate data for the second task, while the entire script and all retrieved products are used for the third subtask.

\begin{table}[t]
\small
\centering
\resizebox{!}{!}{
\begin{tabular}{@{}l|lll@{}}
\toprule
Type & \#Data (Unlabeled) & \#Token & Expert.  \\ 
\midrule
Scripts & 605,229 & 71.5 & 94.0\% \\
Steps & 5,928,271 & 7.48 & 94.0\% \\
\hspace{1mm} w. products & 3,018,276 & 6.98 & - \\
\hspace{1mm} w.o. products & 2,909,995 & 7.98 & -\\
Products & 2,401,087 & 19.31 & - \\
Intentions & 24,010,870 & 10.27 & 98.5\% \\
\midrule
Task 1 & 5,000 (592,729) & - & 95.5\% \\
Task 2 & 5,000 (5,919,278) & - & 96.5\% \\
Task 3 & 5,000 (589,801) & - & 97.0\% \\ 
\bottomrule
\end{tabular}
}
% \vspace{-0.1in}
\caption{Statistics of the~\dataset{} benchmark. \#Token refers to average number of tokens used. Expert. refers to expert acceptance rate.}
% \vspace{-0.25in}
\label{tab:benchmark_statistics}
\end{table}

\subsection{Human Annotations}
\noindent\textbf{Benchmark Annotation:}
We finally conduct human annotations via Amazon Mechanical Turk (AMT) to provide gold labels for a sampled proportion of data and build them into an evaluation benchmark. 
For each task, 5,000 data entries are randomly sampled for annotation. 
We qualify 56 (18.67\%) workers from a pool of 300 candidates with excellent annotation records and provide them with detailed instructions to complete each subtask. 
They are then tasked with annotating (1) the plausibility and feasibility of a given script towards an objective as generated in~\S\ref{sec:user_objective_collection}, (2) the necessity of purchasing a given product for a specific step in the script, as collected in~\S\ref{sec:step_intention_alignment}, and (3) the overall feasibility of a product-enriched script given the user objective, the entire script, and all retrieved products.
Note that only scripts that passed the plausibility annotation are used as candidate data in further tasks. 
We collect five votes for each entry, and the majority vote is used as the final label. 
The overall inter-annotator agreement (IAA) is 78\% in terms of pairwise agreement, and the Fleiss Kappa~\cite{fleiss1971measuring} is 0.53, indicating sufficient agreement. 
More details are in Appendix~\ref{appendix:annotation_details}.

\noindent\textbf{Expert Verification:}
To verify the quality of our collected labels, we invite three additional experts in e-commerce NLP to perform an extra round of annotation verification. 
Each expert is asked to annotate a sample of 200 data entries for each task, following the same instructions provided to the AMT annotators. 
Results in Table~\ref{tab:benchmark_statistics} show that, on average, 96.33\% of the labels collected from AMT annotations align with the experts' majority vote, demonstrating the reliability of our collected labels.

\section{Experiments and Analyses}
\subsection{\dataset{} Statistics}
\begin{figure}[t]
     \centering
     \includegraphics[width=\linewidth]{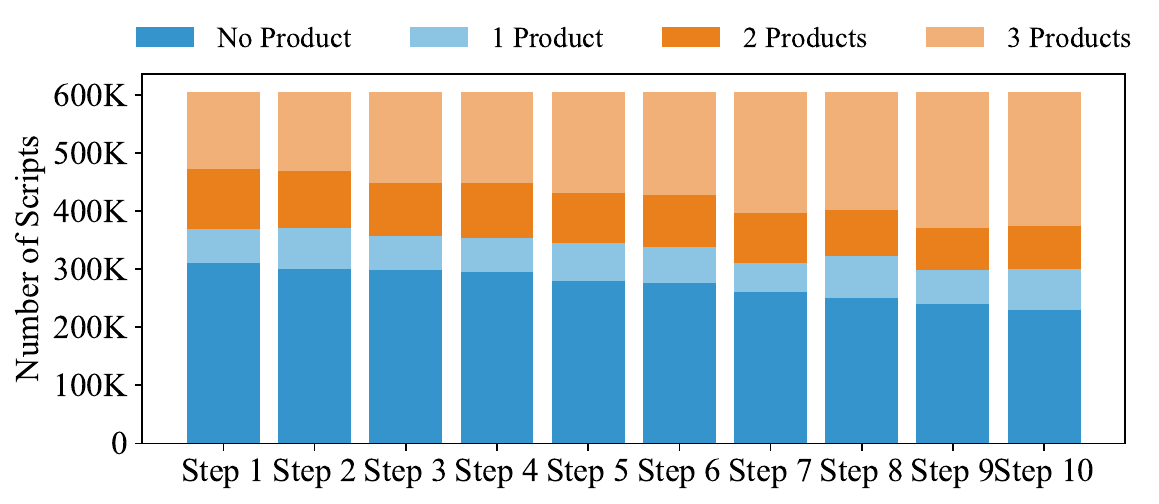}
     % \vspace{-0.3in}
     \caption{Distribution of the number of retrieved products at each step in~\dataset{}.}
    \label{fig:statistics_barplot}
    % \vspace{-0.3in}
\end{figure}

We first present the statistics of \dataset{} in Table~\ref{tab:benchmark_statistics}. 
In total, we collect 605K scripts with 5.9 million steps.
Of a sample of 200 scripts, 94\% were annotated as plausibly correct by expert annotators.
Among them, approximately 3 million steps are deemed necessary for product purchases by the LLM, while the others do not necessitate products. 
We also collect 24 million intentions based on 2.4 million products, of which 98.5\% from a sample of 200 are deemed plausible by expert annotators, demonstrating the high quality of our dataset. 
For each task, we collect labels for 5,000 sampled data entries and leave the rest unlabeled.
We partition the annotated data into train, dev, and test sets according to an 8:1:1 ratio.
Each task follows a high expert acceptance rate, demonstrating the reliability of \dataset{}.
We further visualize the product distributions of our collected scripts by steps in Figure~\ref{fig:statistics_barplot}. 
We observe that as the script progresses (the number of steps increases), more steps are required to associate with products, indicating the need for e-commerce script planning in real-world scenarios.

\begin{table*}[t]
    \small
    \centering
    \resizebox{1\linewidth}{!}{
	\begin{tabular}{@{}llccccccccc@{}}
	\toprule
    \multirow{2}{*}{\textbf{Methods}}&\multirow{2}{*}{\textbf{Backbone}}&\multicolumn{3}{c}{\textbf{Script Verification}} &\multicolumn{3}{c}{\textbf{Product Discrimination}}&\multicolumn{3}{c}{\textbf{Product-Script Veri.}}\\
    \cmidrule(lr){3-5}\cmidrule(lr){6-8}\cmidrule(lr){9-11}
	&&\textbf{Acc}&\textbf{AUC}&\textbf{Ma-F1}&\textbf{Acc}&\textbf{AUC}&\textbf{Ma-F1}&\textbf{Acc}&\textbf{AUC}&\textbf{Ma-F1}\\
            \midrule
            \textbf{Random} & \multicolumn{1}{l}{N/A} & 50.00 & - & 50.00 & 50.00 & - & 50.00 & 50.00 & - & 50.00 \\
            \textbf{Majority} & \multicolumn{1}{l}{N/A} & 60.98 & - & 60.05 & 57.67 & - & 57.10 & 56.46 & - & 56.24 \\
		  \midrule
            \multirow{4}{*}{\textbf{\begin{tabular}[c]{@{}l@{}}PTLM\\ \textit{(Zero-shot)}\end{tabular}}}
            % &RoBERTa-Base \scriptsize{\textit{211M}} & 46.57 & 46.77 & 46.42 & 42.86 & 42.68 & 42.33 & 47.82 & 48.13 & 47.31 \\
            &RoBERTa-Large \scriptsize{\textit{340M}} & 52.04 & 51.79 & 51.21 & 50.80 & 50.74 & 50.68 & 51.39 & 51.37 & 51.32 \\
            % &DeBERTa-Base \scriptsize{\textit{214M}} & 47.61 & 47.77 & 47.28 & 50.94 & 51.02 & 50.05 & 50.34 & 49.99 & 50.27 \\
            &DeBERTa-Large \scriptsize{\textit{435M}} & 51.98 & 52.06 & 51.82 & 52.00 & 51.96 & 51.23 & 52.34 & 52.59 & 51.81 \\
            &CAR \scriptsize{\textit{435M}} & 52.77 & 52.75 & 51.95 & 51.98 & 52.10 & 51.88 & 53.06 & 53.25 & 52.90 \\
            &CANDLE \scriptsize{\textit{435M}} & 53.76 & 53.61 & 53.20 & 52.89 & 53.10 & 52.28 & 52.40 & 52.37 & 51.91 \\
            &VERA-xl \scriptsize{\textit{3B}} & 53.63 & 53.50 & 53.18 & 52.94 & 52.87 & 52.21 & 52.18 & 52.09 & 51.94 \\
            &VERA-xxl \scriptsize{\textit{11B}} & \underline{55.77} & \underline{55.66} & \underline{54.79} & \underline{54.49} & \underline{54.61} & \underline{53.92} & \underline{54.90} & \underline{54.94} & \underline{54.34} \\
            \midrule
            \multirow{13}{*}{\textbf{\begin{tabular}[c]{@{}l@{}}LLM\\ \textit{(Zero-shot)}\end{tabular}}}
            & Meta-Llama-3-8B & 70.05 & - & 69.98 & 64.83 & - & 64.36 & 61.16 & - & 60.22 \\
            & Meta-Llama-3-70B & 71.74 & - & 71.52 & 66.02 & - & 65.05 & 62.00 & - & 61.33 \\
            & Meta-Llama-3.1-8B & 71.45 & - & 71.30 & 65.74 & - & 65.69 & 61.63 & - & 60.96 \\
            & Meta-Llama-3.1-70B & 72.65 & - & 72.42 & 66.15 & - & 65.54 & 62.50 & - & 62.22 \\
            & Meta-Llama-3.1-405B & \underline{75.26} & - & \underline{74.97} & \underline{68.16} & - & \underline{67.33} & \underline{65.66} & - & \underline{65.65} \\
            & Gemma-2-2B & 66.82 & - & 66.80 & 60.56 & - & 60.22 & 58.95 & - & 58.10 \\
            & Gemma-2-9B & 71.27 & - & 70.98 & 65.14 & - & 64.15 & 61.07 & - & 60.40 \\
            & Gemma-2-27B & 71.77 & - & 71.27 & 66.86 & - & 66.20 & 63.15 & - & 62.70 \\
            & Phi-3.5-mini \scriptsize{\textit{4B}} & 68.18 & - & 68.05 & 61.92 & - & 61.15 & 60.36 & - & 59.79 \\
            & Falcon2 \scriptsize{\textit{11B}} & 71.73 & - & 71.68 & 65.70 & - & 65.12 & 61.89 & - & 61.65 \\
            & Mistral-7B-v0.3 & 72.38 & - & 71.49 & 66.42 & - & 65.77 & 62.18 & - & 61.47 \\
            & Mistral-Nemo \scriptsize{\textit{12B}} & 73.18 & - & 72.51 & 66.98 & - & 66.78 & 62.95 & - & 62.71 \\
            & Mixtral-8x7B-v0.1 & 75.06 & - & 74.25 & 66.39 & - & 65.59 & 63.64 & - & 62.84 \\
            \midrule
            \multirow{7}{*}{\textbf{\begin{tabular}[c]{@{}l@{}}PTLM \& LLM\\ \textit{(Fine-tuned)}\end{tabular}}}
            & RoBERTa-Large \scriptsize{\textit{340M}} & 79.18 & 79.27 & 78.86 & 72.26 & 72.32 & 71.74 & 70.26 & 70.38 & 69.83 \\
            & DeBERTa-v3-Large \scriptsize{\textit{435M}} & 81.10 & 80.76 & 81.03 & 74.26 & 74.56 & 73.78 & 72.00 & 71.93 & 71.99 \\
            & Meta-LLaMa-3-8B & 83.48 & 83.38 & 82.64 & 75.75 & 75.52 & \textbf{\underline{75.73}} & 73.06 & 73.33 & 72.84 \\
            & Meta-LLaMa-3.1-8B & 85.24 & 85.07 & 84.64 & \textbf{\underline{76.44}} & \textbf{\underline{76.51}} & 75.53 & \textbf{\underline{74.48}} & \textbf{\underline{74.44}} & \textbf{\underline{74.38}} \\
            & Gemma-2-2B & 81.06 & 80.95 & 80.82 & 73.43 & 73.51 & 73.09 & 69.61 & 69.79 & 68.78 \\
            & Gemma-2-9B & 82.04 & 82.20 & 81.35 & 73.58 & 73.94 & 73.15 & 71.65 & 71.41 & 71.44 \\
            & Mistral-7B-v0.3 & \textbf{\underline{85.72}} & \textbf{\underline{85.61}} & \textbf{\underline{85.51}} & 75.63 & 75.61 & 75.33 & 73.18 & 73.09 & 72.62 \\
            \midrule
            \multirow{10}{*}{\textbf{\begin{tabular}[c]{@{}l@{}}LLM\\ \textit{(API)}\end{tabular}}}
            &GPT4o-mini & 74.30 & - & 73.54 & 69.03 & - & 68.47 & 69.68 & - & 69.16 \\
            &GPT4o-mini (5-shots) & 74.56 & - & 73.61 & 71.56 & - & 71.09 & 71.39 & - & 71.04 \\
            &GPT4o-mini (COT) & 71.66 & - & 71.59 & 69.31 & - & 68.63 & 70.62 & - & 70.23 \\
            &GPT4o-mini (SC-COT) & 72.74 & - & 72.38 & 71.13 & - & 70.79 & 70.93 & - & 70.26 \\
            &GPT4o-mini (SR) & 73.32 & - & 72.35 & 72.46 & - & 71.89 & 71.08 & - & 70.43 \\
            &GPT4o & 77.50 & - & \underline{77.23} & 73.04 & - & 72.06 & 71.50 & - & 71.33 \\
            &GPT4o (5-shots) & \underline{77.92} & - & 76.93 & \underline{73.90} & - & \underline{73.68} & \underline{72.85} & - & \underline{72.83} \\
            &GPT4o (COT) & 74.89 & - & 74.12 & 71.05 & - & 70.58 & 70.32 & - & 69.68 \\
            &GPT4o (SC-COT) & 73.84 & - & 73.16 & 71.08 & - & 70.67 & 69.26 & - & 68.67 \\
            &GPT4o (SR) & 76.22 & - & 76.13 & 71.97 & - & 71.28 & 71.90 & - & 70.96 \\
		\bottomrule
	\end{tabular}
    }
% \vspace{-0.07in}
\caption{Evaluation results (\%) of various (L)LMs on the annotated testing sets of~\dataset{}. 
The best performances within each method are \underline{underlined} and the best among all methods are \textbf{bold-faced}.} 
\label{tab:main_evaluation_results}
\vspace{-0.1in}
\end{table*}

\subsection{Benchmarking Experiments}
\paragraph{Setup:}
We experiment with a selection of (L)LMs to investigate their performance on our proposed tasks. 
Each task, as defined in~\S\ref{sec:task_definitions}, is evaluated as a binary classification task using accuracy, AUC, and Macro-F1 scores as evaluation metrics. 
The evaluation of different models is categorized into three types:
\textbf{(1) \textsc{Zero-shot}:} 
We first evaluate several (L)LMs in a zero-shot manner on the full annotated testing set. 
For small-sized Pre-Trained Language Models (PTLMs), we assess RoBERTa~\cite{RoBERTa}, DeBERTa-v3~\cite{DeBERTav3}, CAR~\cite{CAR}, CANDLE~\cite{CANDLE}, and VERA~\cite{VERA} following the zero-shot question answering evaluation paradigm~\cite{DBLP:conf/aaai/MaIFBNO21}. 
For LLMs, we evaluate Llama3, Llama3.1~\cite{LLAMA2,LLAMA3}, Gemma2~\cite{Gemma,Gemma2}, Phi3.5~\cite{Phi35}, Falcon2~\cite{Falcon2}, Mistral~\cite{Mistral}, and Mixtral~\cite{Mixtral} via direct zero-shot prompting~\cite{DBLP:conf/emnlp/QinZ0CYY23}.
\textbf{(2) \textsc{Finetuning}:}
Next, we assess the performance of LLMs when fine-tuned on \dataset{}. 
We fine-tune RoBERTa, DeBERTa, Llama3, Llama3.1, Gemma2, Falcon2, and Mistral and evaluate them on the partitioned testing set.
LLMs are fine-tuned using LoRA~\cite{Lora}. 
\textbf{(3) \textsc{LLM API}:} 
Finally, we evaluate the performance of GPT4o~\cite{GPT4,GPT4o} and GPT4o-mini~\cite{GPT4omini}, which represent proprietary LLMs, using zero-shot, few-shot~\cite{FewShot}, Chain-of-Thought (COT;~\citealp{COT}), Self-Consistent COT (SC-COT;~\citealp{SCCOT}), and Self-Reflection (SR;~\citealp{DBLP:conf/nips/ShinnCGNY23}) promptings on the full annotated testing set. 
We also include Random and Majority voting to illustrate the characteristics of our benchmark.
See more details in Appendix~\ref{appendix:experiment_implementations}.

\begin{table*}[t]
\small
\centering
\resizebox{1\linewidth}{!}{
\begin{tabular}{@{}ll|ccccccccc@{}}
\toprule
    \multirow{2}{*}{\textbf{Backbone}}&\multirow{2}{*}{\textbf{Training Data}}&\multicolumn{3}{c}{\textbf{Script Verification}} &\multicolumn{3}{c}{\textbf{Product Discrimination}}&\multicolumn{3}{c}{\textbf{Product-Script Veri.}}\\
    \cmidrule(lr){3-5}\cmidrule(lr){6-8}\cmidrule(lr){9-11}
	&&\textbf{Acc}&\textbf{AUC}&\textbf{Ma-F1}&\textbf{Acc}&\textbf{AUC}&\textbf{Ma-F1}&\textbf{Acc}&\textbf{AUC}&\textbf{Ma-F1}\\
\midrule
\multirow{4}{*}{\textbf{\begin{tabular}[c]{@{}l@{}}Llama-3.1\\ \scriptsize{\textit{8B}}\end{tabular}}} & Zero-shot & 71.45 & - & 71.30 & 65.74 & - & 65.69 & 61.63 & - & 60.96 \\
& \dataset{} & 83.86 & 83.94 & 83.05 & 77.70 & 77.87 & 77.59 & 75.88 & 75.58 & 75.58 \\
& FolkScope + MIND & 67.74 & 67.63 & 67.38 & 66.79 & 66.43 & 66.11 & 64.91 & 64.87 & 64.42 \\
& \hspace{1mm} + \dataset{} & \underline{84.65} & \underline{84.84} & \underline{84.13} & \underline{78.60} & \underline{78.83} & \underline{78.27} & \underline{76.35} & \underline{76.50} & \underline{76.08} \\
\midrule
\multirow{4}{*}{\textbf{\begin{tabular}[c]{@{}l@{}}Mistral-v0.3\\ \scriptsize{\textit{7B}}\end{tabular}}} & Zero-shot & 72.38 & - & 71.49 & 66.42 & - & 65.77 & 62.18 & - & 61.47 \\
& \dataset{} & 85.72 & 85.61 & 85.51 & 75.63 & 75.61 & 75.33 & 73.18 & 73.09 & 72.62 \\
& FolkScope + MIND & 69.77 & 70.00 & 69.56 & 67.78 & 67.75 & 67.39 & 63.70 & 63.41 & 63.66 \\
& \hspace{1mm} + \dataset{} & \textbf{\underline{85.87}} & \textbf{\underline{85.80}} & \textbf{\underline{86.37}} & \textbf{\underline{81.18}} & \textbf{\underline{80.96}} & \textbf{\underline{80.54}} & \textbf{\underline{78.94}} & \textbf{\underline{78.94}} & \textbf{\underline{78.66}} \\
\bottomrule
\end{tabular}
}
% \vspace{-0.1in}
\caption{Evaluation results (\%) of transfering knowledge from FolkScope and MIND to aid~\dataset{}.
The best performances among each method is \underline{underlined} and best ones among all methods are \textbf{bold-faced}.
}
\label{tab:transfer_FolkScope}
\vspace{-0.1in}
\end{table*}

\paragraph{Results:}
The evaluation results are presented in Table~\ref{tab:main_evaluation_results}. 
Our observations include:
\textbf{(1) Challenges with~\task{} Tasks:} (L)LMs struggle with all tasks in e-commerce script planning, particularly in those involving e-commerce products. 
All models achieved only moderately satisfactory performance across the three subtasks. 
For instance, the best open-source LLM, LLAMA-3.1-405B, attained accuracy scores of 75\%, 68\%, and 65\% on the respective tasks. 
This underscores the inherent difficulty of the \dataset{}. 
Notably, the latter two subtasks are considerably more challenging than script verification, likely due to the complexities associated with e-commerce products and the requisite product knowledge.
\textbf{(2) Impact of Fine-tuning and Advanced Prompting:} While fine-tuning and advanced prompting methods yield some performance improvements, there remains significant room for enhancement. 
We observed a notable boost in performance when LLMs are fine-tuned on annotated product-enriched scripts. 
For example, the performance of LLAMA-3.1-8B improved by 12\%, 11\%, and 13\% across the three tasks, respectively. 
Similarly, GPT series models benefited from advanced prompting techniques, such as few-shot prompting and self-reflection.
COT prompting, on the other hand, cannot help, which may be due to its reliance on the model’s internal reasoning paths rather than incorporating additional external product-related signals or domain-specific annotations that align closely with the given tasks.
\textbf{(3) Effects of Model Training Paradigms and Scale:} Enhancing the training paradigm and increasing the number of parameters positively impacted performance. 
In the LLAMA series, both increasing parameters and updating training data and methods led to improved results. 
The performance trend associated with increasing the number of parameters is also clear and highlights the significance of model scale in achieving better outcomes on our tasks.
\textbf{(4) Complexity of Tasks:} The poor performances on both the step-product discrimination and script-product verification tasks demonstrate that \task{} is a complex and challenging problem for LLMs, revealing the limitations of current models in flexibly integrating e-commerce product knowledge into planning tasks.
Greater efforts should be directed along this direction in order to achieve automated e-commerce script planning in a single-step generative manner.
% More analyses are in Appendix~\ref{appendix:additional_experiments}.

\subsection{The Effect of Injecting Intentions}
From the evaluation results in Table~\ref{tab:main_evaluation_results}, we observe that a key weakness in current LLMs is their difficulty in associating products with each step in a script and verifying whether the entire script can work. 
To improve this, we hypothesize that injecting intentional knowledge into LLMs may help, as it provides a better understanding of what e-commerce products can help or how they can assist the customer, thereby promoting the linking of products with script planning.
To achieve this, we select two intention knowledge bases based on products from Amazon, FolkScope~\cite{FolkScope}, and MIND~\cite{MIND} as sources of intentions. 
We use a natural language prompt to concatenate product metadata (title, features, descriptions) as the input with their purchase intentions as the output, and train LLMs under a generative objective using LoRA~\cite{Lora}.
They are then sequentially fine-tuned on training set of~\dataset{}. 
Another group of LLMs, after fine-tuning on FolkScope and MIND, is directly evaluated for comparison. 
All models are evaluated on the testing set of~\dataset{}, and the results are reported in Table~\ref{tab:transfer_FolkScope}.
From the results, we observe a significant improvement across all tasks when the models are sequentially fine-tuned on FolkScope and MIND, then on~\dataset{}, compared to being solely fine-tuned on either one. 
This indicates that aligning LLMs with more e-commerce products' use cases or purchase motivations enhances their ability to identify useful products for users' desired actions or steps in scripts. 
Since intentions from both resources are distilled from LLMs, this opens up a scalable yet cost-efficient paradigm for improving LLMs' performance on e-commerce script planning tasks.

\subsection{Error Analysis of GPT-Series Models}
Finally, for a more fine-grained error analysis, we manually inspect the causes of errors in 200 sampled COT responses generated by GPT-4o across all tasks and categorize their mistakes into three categories:
\textbf{(1) Wrong understanding of products:} 68\% of errors are caused by the LLM's false understanding of a specific usage or feature of a product that conflicts with the steps or the entire script. 
For example, when a step requires controlling the user's non-compatible smart light bulbs using the virtual assistant, the LLM might incorrectly suggest voice commands that only work for compatible devices. 
To address this issue, multi-modal product images or more detailed attributes can be incorporated.
\textbf{(2) Conflict in reasoning across steps:} 27\% of errors occur due to the model's failure to reason about the feasibility of collaborating products associated with different steps, where the model may mistakenly deem it infeasible to purchase two products simultaneously.
\textbf{(3) Internal conflict and annotation errors:} 5\% of errors are due to internal conflicts, such as inconsistencies between the binary predictions made and the corresponding reasoning rationales, as well as annotation errors, potentially caused by overzealous annotators.

\subsection{Category-wise Performance Analysis}
\label{appendix:performance_category}
We then conduct a detailed analysis of GPT-4o’s performance in the product discrimination task across a variety of product categories. 
Table~\ref{tab:accuracy_by_category} presents the accuracy scores obtained for each major product category.
We observe that GPT-4o performs best in categories like ``Toys and Games,'' ``Patio Lawn and Garden,'' ``Grocery and Gourmet Food,'' and ``Cell Phones and Accessories,'' often surpassing 80\% accuracy. 
These categories tend to have clearer, more distinct product descriptors, making it easier to distinguish between items.
In contrast, performance dips in more ambiguous or heterogeneous categories like ``Beauty and Personal Care'' and ``Health and Household.'' Products in these domains often share overlapping descriptors or subtle differences (e.g., similar lotions or vitamins), making text-only differentiation challenging.
Intermediate results in categories like ``Electronics and Office Products'' suggest that while technical specifications are helpful, the sheer diversity of items can still obscure product distinctions. 
Integrating additional modalities, such as images or structured product metadata, might help address these difficulties and improve the model’s discrimination capabilities across a broader range of categories.

\begin{table}[t]
\small
\centering
\begin{tabular}{l|c}
\toprule
Category & Accuracy \\ 
\midrule
Automotive & 64.58 \\
Beauty and Personal Care & 63.95 \\
Cell Phones and Accessories & 82.31 \\
Clothing Shoes and Jewelry & 78.99 \\
Electronics & 66.15 \\
Health and Household & 62.08 \\
Home and Kitchen & 65.63 \\
Grocery and Gourmet Food & 82.49 \\
Industrial and Scientific & 79.51 \\
Office Products & 67.42 \\
Patio Lawn and Garden & 82.84 \\
Sports and Outdoors & 76.68 \\
Tools and Home Improvement & 65.57 \\
Toys and Games & 84.37 \\
\bottomrule
\end{tabular}
\caption{Accuracy (\%) of GPT-4o on product discrimination task by product categories.}
\vspace{-0.1in}
\label{tab:accuracy_by_category}
\end{table}

\section{Conclusions}
In conclusion, this paper proposes the task of e-commerce script planning and introduces a novel framework for collecting product-enriched scripts. 
By applying the framework to Amazon product data, we construct a sibling large-scale knowledge base and build the very first evaluation benchmark upon it. 
Extensive experiments demonstrate the challenges of our task and potential solutions to improve the performance of LLMs on~\task{}. 
We hope that our task and benchmark can serve as an important cornerstone to advance the e-commerce shopping experience by creating more intelligent and personalized shopping assistants with e-commerce script planning capability that ultimately benefit the community and the world.

\newpage
\section*{Limitations}
We discuss three main limitations of our work.

First, our data construction process relies significantly on GPT-4o-mini, a proprietary LLM, for data collection, as well as human annotation for label collection and verification. 
This raises concerns about the reproducibility and the high costs associated with our dataset. 
However, the expense of using GPT-4o-mini is relatively low compared to other proprietary LLMs; for instance, we spent only around \$250 USD to collect 24 million intentions. 
The quality of the output remains outstanding, with fast generation speeds that effectively simulate a real-world LLM-powered shopping assistant. 
We also experimented with using LLAMA-3.1-405B as the core generator for data collection, which also yields exceptional data quality. 
However, hosting the model and using it for inference proved to be computationally and time-intensive, leading us to ultimately choose GPT-4o-mini.

Next, we assign the verification of product compatibility between different steps to a human-annotated task and do not implement any strategies within our data collection framework. 
This decision is made because detecting conflicting products is a complicated task that requires consideration of many features, some of which cannot be determined solely based on product metadata. 
We leave this verification to future industrial efforts to ensure that products retrieved at different stages can accommodate each other and collectively contribute to successful execution.

Finally, we defer the exploration of practical solutions to assist LLMs in solving~\task{}, as well as the deployment of these solutions to deliver real-world benefits, to future work.
We can also implement knowledge editing techniques to address this, as done by~\citet{DBLP:journals/corr/abs-2410-14276,DBLP:journals/corr/abs-2412-11418}.
In the long run, we envision a model capable of accurately understanding a customer’s needs and recommend all products at once via e-commerce script planning can promote purchase decision-making and increase e-commerce revenue.

\section*{Ethics Statement}
Since our dataset curation pipeline involves prompting LLMs, it is important to implement stringent measures to ensure the absence of offensive content in both the prompts and the generated responses. 
We first explicitly state in the prompt that the LLM should not generate any content that contains personal privacy violations, promotes violence, racial discrimination, hate speech, sexual content, or self-harm. 
Then, we manually inspect a random sample of 500 data entries from all tasks in~\dataset{} for offensive content. 
Based on our observations, we have not detected any offensive content. 
Therefore, we believe that our dataset is safe and will not yield any negative societal impact.

Due to data privacy issues, our dataset will not be made public.
% We will share our data and models under the MIT license, thereby granting other researchers free access to our assets for research purposes. 
% If there are any issues with rights violations or anything related to the data license, we will take full responsibility. 
% The Amazon Review Dataset used in this paper is shared under the CC-SA license, permitting us to use it for research. 
As for language models, we access all open-source LMs via the Hugging Face Hub~\cite{DBLP:conf/emnlp/WolfDSCDMCRLFDS20}. 
The number of parameters is presented in Table~\ref{tab:main_evaluation_results}. 
All associated licenses permit user access for research purposes, and we have agreed to follow all terms of use.

We conduct large-scale human annotations on the Amazon Mechanical Turk (AMT) platform. 
We invite annotation workers from the US, Europe, and India due to their proficiency in English. 
The annotators are paid an average hourly rate of \$17.50, which is comparable to the minimum wage in their local jurisdictions. 
The selection of these annotators is solely based on their performance on the evaluation set, and we do not collect any personal information about the participants from AMT. 
The expert annotators agree to participate as their contribution to the paper without compensation. 

\section*{Acknowledgements}
We thank the anonymous reviewers and the area chair for their constructive comments. 
The authors of this paper were supported by the ITSP Platform Research Project (ITS/189/23FP) from the ITC of Hong Kong, SAR, China, as well as the AoE (AoE/E-601/24-N), the RIF (R6021-20), and the GRF (16205322) from the RGC of Hong Kong, SAR, China.
We also thank the Amazon Search Experience Science team for supporting this intern project. 

\bibliography{custom}

\newpage
\appendix
\begin{center}
    {\Large\textbf{Appendices}}
\end{center}

\section{Implementation Details}
\label{appendix:implementation_details}
\subsection{Dataset Construction Prompts}
\label{appendix:dataset_construction_prompt}
We first present the prompts used in each step to sequentially instruct GPT-4o-mini to generate candidate data for \dataset{}. 
Special tokens, such as \texttt{<example>, </example>}, are added whenever necessary throughout the prompts.

\subsubsection{User Objective and Script Collection}
To collect real-world user objectives and their associated scripts, we use the following prompt to instruct the LLM.
\begin{displayquote}
\textbf{\texttt{\small 
Given a product and an user review about it, infer some potential action goals of purchasing the product. 
The goals should be what the user wants to do with the product.
They do not have to be explicitly stated in the review, but can be reasoned from the context. You may think of big goals of what the user wants to achieve with the help of the product.
Action goals should be specific and actionable objectives that take multiple steps to achieve, and the product may contribute to one step of them.
Do not generate goals that are too simple. Do not generate buying a product again, recommend the product or brand to others, reliable customer service, etc.
They shouldn't be very long-term goals, do not generate being successful or making a lot of money.
Separate each goal with || and make each goal specific by describing it in detail. Follow these examples:
}}\\
\ldots \\
\textbf{\textcolor{headcolor}{\texttt{\small Product <i>: Nike Air Zoom Running Shoes}}}\\
\textbf{\textcolor{headcolor}{\texttt{\small Review <i>: This is the best pair of running shoes I've ever owned. They are comfortable and provide great support.}}}\\
\textcolor{tailcolor}{\texttt{\small\textbf{Goals: participate in a marathon || stay a healthy lifestyle || start running regularly}}}\\
\ldots \\
\textbf{\textcolor{headcolor}{\texttt{\small Product <N>: Samsung 65-Inch 4K Smart TV}}}\\
\textbf{\textcolor{headcolor}{\texttt{\small Review <N>: This TV has great picture quality and sound. It's perfect for watching movies and shows.}}}\\
\end{displayquote}

We drop reviews that are less than 10 tokens or contain fewer than 3 unique tokens. 
Additionally, we exclude reviews with more than 5 hashtags, as these are sometimes misleading. 
These thresholds were determined based on our prior experience in processing e-commerce reviews and have proven to provide the best trade-off in retaining the maximum number of valid reviews.
When reviews are of poor quality or unavailable, our offline experiments show that LLMs can still infer potential user objectives from the product title and metadata alone. 
This ensures that the framework remains functional and capable of reasoning about product use cases, even without relying on user reviews.

We then use the following prompt to generate a goal-oriented script based on the objectives collected above. 
A brief explanation of each step is required for clarification.
\begin{displayquote}
\textbf{\texttt{\small 
Given an actionable goal, generate a script of steps that can be used to achieve the goal.
The script should be detailed and specific, and each step should be actionable and achievable sequentially.
Limit the script to within 10 steps and each step to within 20 words. For each step, a short explanation of the step should be provided.
The steps should be in the correct logical and temporal order and should be detailed enough to be executed sequentially by someone who is not familiar with the goal. 
For each step, try to ensure that some E-commerce products can be purchased to help achieve the step.
}}\\
\ldots \\
\textbf{\textcolor{headcolor}{\texttt{\small Objective <i>: Participate in a marathon.}}\\
\textcolor{tailcolor}{\texttt{\small\textbf{Step 1: Choose a marathon to participate in (Research and select a marathon that fits your schedule and location preference)}}}\\
\textcolor{tailcolor}{\texttt{\small\textbf{Step 2: Register for the marathon (Complete the registration form and pay any associated fees)}}}\\
\textcolor{tailcolor}{\texttt{\small\textbf{Step 3: Create a training plan (Develop a schedule with incremental mileage increases and rest days)}}}\\
\textcolor{tailcolor}{\texttt{\small\textbf{Step 4: Purchase proper running gear (Buy running shoes, moisture-wicking clothing, and a water bottle)}}}\\
\textcolor{tailcolor}{\texttt{\small\textbf{Step 5: Start your training program (Follow your schedule, gradually increasing your running distance each week)}}}\\
\textcolor{tailcolor}{\texttt{\small\textbf{Step 6: Maintain a balanced diet (Eat a mix of carbohydrates, proteins, and fats to fuel your training)}}}\\
\textcolor{tailcolor}{\texttt{\small\textbf{Step 7: Stay hydrated (Drink plenty of water daily and during your runs)}}}\\
\textcolor{tailcolor}{\texttt{\small\textbf{Step 8: Practice long runs (Include one long run per week to build endurance, following your training plan)}}}\\
\textcolor{tailcolor}{\texttt{\small\textbf{Step 9: Get enough rest (Ensure adequate sleep and recovery time to avoid overtraining and injuries)}}}\\
\textcolor{tailcolor}{\texttt{\small\textbf{Step 10: Plan race day logistics (Prepare your transportation, know the race course, and plan post-race recovery)}}}
}\\
\ldots \\
\textbf{\textcolor{headcolor}{\texttt{\small Objective <N>: Learn to play the guitar.}}}
\end{displayquote}
For scripts that are more than 10, we drop the outlier steps. 
In total, we recorded only 3,098 cases where truncation occurs, which is very rare.
In most cases, LLM follows our instruction precisely.

\subsubsection{Purchase Intention Mining}
We then distill product purchase intentions from the LLM by following \citet{FolkScope,COSMO} with the prompt below. 
This type of intention distillation has been proven effective and can support downstream applications. 
Thus, our knowledge distillation-based method is justifiable and enables large-scale benchmark construction.
\begin{displayquote}
\textbf{\texttt{\small 
Given a product retrieved from Amazon, you are required to generate 10 possible intentions that a user may have that motivates them to purchase the product.
The intention should be describing what the user wants to do with the product, believe the product can help them achieve, or the problem the product can solve.
It should be specific and not too general. For example, "like the product", "wants to buy it", "good product" are invalid intentions.
Best intentions describe the user's goal, desire, or action to be taken with the product.
Generate the intentions directly with one intention per line. Follow these examples:
}}\\
\ldots \\
\textbf{\textcolor{headcolor}{\texttt{\small Product <i>: Nike Air Zoom Running Shoes}}\\
\textcolor{tailcolor}{\texttt{\small\textbf{Intention 1: Improve their running performance with better cushioning and support.}}}\\
\textcolor{tailcolor}{\texttt{\small\textbf{Intention 2: Train for a marathon or other long-distance race.}}}\\
\textcolor{tailcolor}{\texttt{\small\textbf{Intention 3: Increase comfort during daily jogging sessions.}}}\\
\textcolor{tailcolor}{\texttt{\small\textbf{Intention 4: Reduce the risk of injuries by using shoes with advanced technology.}}}\\
\textcolor{tailcolor}{\texttt{\small\textbf{Intention 5: Enhance their athletic appearance with stylish and modern footwear.}}}\\
\textcolor{tailcolor}{\texttt{\small\textbf{Intention 6: Replace worn-out running shoes with a high-quality, durable option.}}}\\
\textcolor{tailcolor}{\texttt{\small\textbf{Intention 7: Experience the benefits of lightweight shoes for faster running times.}}}\\
\textcolor{tailcolor}{\texttt{\small\textbf{Intention 8: Participate in a running club or group with appropriate gear.}}}\\
\textcolor{tailcolor}{\texttt{\small\textbf{Intention 9: Transition to a more serious and dedicated running routine.}}}\\
\textcolor{tailcolor}{\texttt{\small\textbf{Intention 10: Alleviate foot pain caused by inadequate or poorly fitting shoes.}}}
}\\
\ldots \\
\textbf{\textcolor{headcolor}{\texttt{\small Product <N>: chouyatou Women's Casual Long Sleeve Button Down Loose Striped Cotton Maxi Shirt Dress}}}\\
\end{displayquote}

Using intention as the connecting link is inherently more effective than traditional search queries, such as keywords in product titles and metadata, which are often not represented in script steps.
To best align intentions with steps, we create exemplars with similar semantics and grammatical structures to effectively guide GPT-4o-mini in generating steps and intentions with consistent linguistic patterns (e.g., omitting the subject, using the simple present tense, and keeping them short and concise). 
However, gaps between intentions and steps can still occur. 
To address this, we generate 10 intentions per product to ensure as much coverage as possible. 
In industrial applications, even more intentions per product could be generated to enhance coverage and improve alignment further, given the low cost of generating outputs with GPT-4o-mini.
In our current dataset construction pipeline, expert evaluations and human annotations confirm that our method is effective and does not significantly impact final performance. 
However, verifying its efficacy at an industrial scale is left to future work by the e-commerce community.

\subsubsection{Step-Intention Alignment}
Finally, we prompt the LLM again to determine whether a product purchase is necessary for each step in the script. For steps that are deemed necessary, we ask the LLM to generate a list of keywords to help us narrow down the search scope of products and proceed with our intention alignment strategy. 
We use the following prompt to assess the purchase necessity of each step:
\begin{displayquote}
\textbf{\texttt{\small 
Given a plan consisting of ten steps, determine whether any additional item or product can be helpful in each step to make it successful. Note that it can be anything or any product that is helpful in terms of achieving the step.
There are steps that definitely do not require additional help from other things or items, such as "inviting a friend", "going to somewhere", "select a time", "search for a specific information". They are usually actions that can be done directly by the person and do not require additional assistance from a product. 
There are also steps that can be assisted by having other products, such as "prepare food", "clean the house", "write a letter", "make a phone call", "prepare entertainment". They usually involve interactions with some tools, materials, or other things to complete the action, or can be done mor easily or efficiently with the help of them.
Given the steps below, first provide a yes or no answer to whether it is helpful to purchase a product to achieve the step. Then, provide a short list of product keywords that represent items that can be helpful in achieving the step. These keywords can be general to represent more items.
You are forced to follow the example format in generating the answer, which is first generate a one word answer, either "yes" or "no", then generate a list of keywords. Generate one line per step. For example:
}}\\
\ldots \\
\textbf{\textcolor{headcolor}{\texttt{\small Step 1: Get measuring cups and spoons (Purchase a set with common baking measurements to accurately measure ingredients)}}\\
\textcolor{headcolor}{\texttt{\small Step 2: Get a food scale (Weigh ingredients like meat and flour for reliable portioning)}}\\
\textcolor{headcolor}{\texttt{\small Step 3: Use a thermometer (Monitor oil and internal temperatures for frying and roasting)}}\\
\textcolor{headcolor}{\texttt{\small Step 4: Time activities (Use a timer to track marinating, baking, etc. for consistency)}}\\
\textcolor{headcolor}{\texttt{\small Step 5: Watch tutorial videos (View cooking demos to learn proper knife skills and techniques)}}\\
\textcolor{headcolor}{\texttt{\small Step 6: Take an in-person cooking class (Learn from a professional chef for hands-on experience)}}\\
\textcolor{headcolor}{\texttt{\small Step 7: Practice fundamental recipes (Master basic recipes to handle ingredients and temperatures)}}\\
\textcolor{headcolor}{\texttt{\small Step 8: Focus on one technique (Work on skills like sautéing, searing, or deglazing)}}\\
\textcolor{headcolor}{\texttt{\small Step 9: Invest in high-quality cookware (Buy pans that distribute heat evenly for optimal cooking)}}\\
\textcolor{headcolor}{\texttt{\small Step 10: Follow recipes precisely (Carefully measure and time each step before improvising)}}\\
\textcolor{tailcolor}{\texttt{\small yes (measuring cups, spoons, measurement, ingredients)}}\\
\textcolor{tailcolor}{\texttt{\small yes (food scale, scale, weight)}}\\
\textcolor{tailcolor}{\texttt{\small yes (thermometer, oil, temperature)}}\\
\textcolor{tailcolor}{\texttt{\small yes (timer, activities, marinating)}}\\
\textcolor{tailcolor}{\texttt{\small yes (tutorial videos, cooking demos, knife skills)}}\\
\textcolor{tailcolor}{\texttt{\small no}}\\
\textcolor{tailcolor}{\texttt{\small yes (fundamental recipes, basic recipes, recipes)}}\\
\textcolor{tailcolor}{\texttt{\small no}}\\
\textcolor{tailcolor}{\texttt{\small yes (high-quality cookware, pans, heat)}}\\
\textcolor{tailcolor}{\texttt{\small yes (recipes, measure, time)}}
}\\
\ldots \\
\textbf{\textcolor{headcolor}{\texttt{\small\textbf{Step 1: Choose a marathon to participate in (Research and select a marathon that fits your schedule and location preference)}}}\\
\textcolor{headcolor}{\texttt{\small\textbf{Step 2: Register for the marathon (Complete the registration form and pay any associated fees)}}}\\
\textcolor{headcolor}{\texttt{\small\textbf{Step 3: Create a training plan (Develop a schedule with incremental mileage increases and rest days)}}}\\
\textcolor{headcolor}{\texttt{\small\textbf{Step 4: Purchase proper running gear (Buy running shoes, moisture-wicking clothing, and a water bottle)}}}\\
\textcolor{headcolor}{\texttt{\small\textbf{Step 5: Start your training program (Follow your schedule, gradually increasing your running distance each week)}}}\\
\textcolor{headcolor}{\texttt{\small\textbf{Step 6: Maintain a balanced diet (Eat a mix of carbohydrates, proteins, and fats to fuel your training)}}}\\
\textcolor{headcolor}{\texttt{\small\textbf{Step 7: Stay hydrated (Drink plenty of water daily and during your runs)}}}\\
\textcolor{headcolor}{\texttt{\small\textbf{Step 8: Practice long runs (Include one long run per week to build endurance, following your training plan)}}}\\
\textcolor{headcolor}{\texttt{\small\textbf{Step 9: Get enough rest (Ensure adequate sleep and recovery time to avoid overtraining and injuries)}}}\\
\textcolor{headcolor}{\texttt{\small\textbf{Step 10: Plan race day logistics (Prepare your transportation, know the race course, and plan post-race recovery)}}}
}
\end{displayquote}

For SentenceBERT, we use T5-xxl (11B; ~\citealp{DBLP:journals/jmlr/RaffelSRLNMZLL20}) as the backbone. 
We begin by calculating the embeddings of all purchase intentions for a product and all steps separately, then compute the semantic similarity between every pair of intention and step using cosine similarity as the metric. 
Each product's purchase relatedness is determined by the average similarity of all pairs relevant to the product for a specific step. 
To ensure that only related products are selected, we set a lower bound threshold of $\tau=0.4$, which filters out approximately 95\% of products at each step, improving the data quality.

Analyzing the distribution of embedding similarity, we find that 13\% of intentions have a similarity score higher than 0.5 with the user objective. 
If we were to eliminate cases where intentions closely match objectives, the effectiveness of product retrieval would likely decrease. 
This is because many product recommendations rely on the semantic alignment between intentions and specific steps in a user's script. 
Removing these closely matched cases could lead to gaps in relevant product associations, resulting in less accurate or relevant recommendations. 
Therefore, we argue that non-script-level intentions—those not directly tied to a specific step—also play a crucial role in improving product retrieval and the overall user experience.

\begin{table*}[t]
\small
\centering
\resizebox{\linewidth}{!}{
\begin{tabular}{@{}l|l@{}}
\toprule
Task & Prompt \\ 
\midrule
SV. & \begin{tabular}[c]{@{}l@{}}You are given an objective \textbf{\texttt{<TEST-ENTRY-OBJECTIVE>}} and a script \textbf{\texttt{<TEST-ENTRY-SCRIPT>}}.\\ Your task is to assess the plausibility and feasibility of the script in relation to the objective. \\First, evaluate the plausibility by determining if the script logically aligns with the objective. \\Next, consider the feasibility by assessing whether the script is realistic and achievable given the constraints of the objective.\\
Based on your evaluation, please output a binary answer ``yes'' or ``no''.\\
``yes'' indicates that the script is both plausible and feasible.
``no'' indicates that the script is either implausible or infeasible. \\
Please answer with one word ``yes'' or ``no'':\end{tabular}\\ 
\midrule
PD. & \begin{tabular}[c]{@{}l@{}}You are given an objective \textbf{\texttt{<TEST-ENTRY-OBJECTIVE>}}, a specific action\\ \textbf{\texttt{<TEST-ENTRY-STEP>}}, and a product \textbf{\texttt{<TEST-ENTRY-PRODUCTT>}}.\\
Your task is to determine whether the step requires the purchase of a product to assist the user in accomplishing that step.\\
First, assess if the step \textbf{\texttt{<TEST-ENTRY-STEP>}} necessitates any product purchase.\\
If it does, evaluate whether purchasing the product \textbf{\texttt{<TEST-ENTRY-PRODUCT>}} can effectively help with the step.\\
Based on your evaluation, please output a binary answer ``yes'' or ``no''.\\
``yes'' indicates that the product is a good match and can contribute to the step.\\
``no'' indicates that the step does not require any product purchase or that the product cannot help.\\
Please answer with one word ``yes'' or ``no'':\end{tabular}\\ 
\midrule
SPV. & \begin{tabular}[c]{@{}l@{}}You are given an objective \textbf{\texttt{<TEST-ENTRY-OBJECTIVE>}}, a script consisting of multiple steps \textbf{\texttt{<TEST-ENTRY-SCRIPT>}},\\ 
and the products associated with each step \textbf{\texttt{<TEST-ENTRY-PRODUCT-ENRICHED-SCRIPT>}}.\\
Your task is to determine the overall feasibility of the product-enriched script.\\
Evaluate whether any internal conflicts exist between the different steps and their associated products.\\
If all products are suitable for their respective steps and can collaborate effectively within the entire script.\\
Based on your evaluation, please output a binary answer ``yes'' or ``no''.\\
``yes'' indicates that all products are appropriate and can work together seamlessly.\\
``no'' indicates that there are internal conflicts among the steps and products.\\
Please answer with one word ``yes'' or ``no'':\end{tabular}\\ 
\bottomrule
\end{tabular}
}
\caption{Evaluation prompts used for benchmarking LLMs' performances across three tasks in ~\dataset{}: SV, PD, and SPV refer to: Script Verification, Product Discrimination, and Script-Product Verification.}
\label{tab:appendix_evaluation_prompts}
\end{table*}

\subsection{Evaluation Prompts}
\label{appendix:evaluation_prompts}
To evaluate LLMs on three tasks in \dataset{}, we present our evaluation prompts in a zero-shot scenario in Table\ref{tab:appendix_evaluation_prompts}.
These prompts are consistently used across all model evaluations to ensure a fair comparison.

For few-shot evaluations, examples are added after the task descriptions and before the prompted test entry. 
The exemplars are randomly sampled for each test entry from a set of 20 expert-annotated examples.

For Chain of Thought (COT) prompting, we specifically instruct LLMs to "think step by step and generate a short rationale to support your reasoning." We then ask them to provide an answer based on the generated rationale. The sampling temperature, $\tau$, is set to 0.1 by default, and 5 COT responses are sampled with $\tau$ set to 0.7 in the SC-COT setting. 
In the SC-COT setting, we also explicitly include another round of conversation to allow the LLM to verify whether the prediction is correct according to its generated rationale.

For self-reflection, we follow previous approaches~\cite{DBLP:conf/acl/WangZ0MZ024,DBLP:conf/www/KoaMNC24} and construct similar prompts to evaluate the LLM.

\subsection{Evaluation Implementations}
\label{appendix:experiment_implementations}
To evaluate PTLMs in a zero-shot manner, we adopt the evaluation pipeline used for zero-shot question answering~\cite{DBLP:conf/aaai/MaIFBNO21,CAR,DBLP:conf/acl/WangFXBSC23,DBLP:journals/corr/abs-2406-10885}.
Specifically, for each task, we convert the question into two declarative statements, which serve as natural language assertions corresponding to `yes'' or ``no'' options. 
For instance, when determining whether a product is necessary for a step, we generate two assertions: ``The product \texttt{<PRODUCT>} is helpful to the step \texttt{<STEP>},'' and ``The product \texttt{<PRODUCT>} is not helpful to the step \texttt{<STEP>}.''
The models are then tasked with computing the loss of each assertion. 
The assertion with the lowest loss is considered as the model's prediction. 
This approach allows any PTLM to be evaluated under classification tasks with an arbitrary number of options or even type classification based on a single assertion. 
We use the open code library\footnote{\href{https://github.com/Mayer123/HyKAS-CSKG}{https://github.com/Mayer123/HyKAS-CSKG}} as our code base and follow the default hyperparameter settings.
For VERA, we follow the exact same implementation\footnote{\href{https://github.com/liujch1998/vera}{https://github.com/liujch1998/vera}}~\cite{VERA}.
The accessed backbone models are \texttt{liujch1998/vera-xl} (3B) and \texttt{liujch1998/vera} (11B), and all other hyperparameter settings follow the default setting.

For evaluating LLMs in a zero-shot manner, we transform the input for each task into assertions using natural language prompts, as explained in Appendix~\ref{appendix:evaluation_prompts} and Table~\ref{tab:appendix_evaluation_prompts}.
The models are then prompted to determine the plausibility of the provided assertions by answering yes or no questions.
We parse their responses using pre-defined rules to derive binary predictions.
When generating each token, we consider the top 10 tokens with the highest probabilities.
Their generation process is limited to 10 tokens for computational efficiency.

For fine-tuning LLMs, we use LoRA for fine-tuning, and the LoRA rank and $\alpha$ are set to 16 and 32, respectively.
We adopt the open code library from LlamaFactory\footnote{\href{https://github.com/hiyouga/LLaMA-Factory}{https://github.com/hiyouga/LLaMA-Factory}}~\cite{LLamaFactory} for model training and evaluation.
We similarly use an Adam~\cite{DBLP:journals/corr/KingmaB14} optimizer with a learning rate of 5e-5 and a batch size of 8.
The maximum sequence length for the tokenizer is set at 300.
All models are fine-tuned over three epochs and the last checkpoint is evaluated.
We use three random seeds and report the average performance for all experiments.

Finally, for evaluating proprietary LLMs, such as GPT-4o and GPT-4o-mini, we similarly prompt them as with open LLMs.
Detailed prompts are explained in Appendix~\ref{appendix:evaluation_prompts}.

\section{Annotation Details}
\label{appendix:annotation_details}
\subsection{Worker Selection Protocol}
To ensure the high quality of our human annotation, we implement strict quality control measures.
Initially, we invite only those workers to participate in our qualification rounds who meet the following criteria: 1) a minimum of 2,000 HITs approved, and 2) an approval rate of at least 90\%. 
We select workers separately for each task and conduct three qualification rounds per task to identify those with satisfactory performance. 
In each qualification round, we create a qualification test suite that includes both easy and challenging questions, each with a gold label from the authors. 
Workers are required to complete a minimum of 40 questions. 
To qualify, they must achieve an accuracy rate of at least 75\% on the qualification test. 
After our selection process, we chose 56 workers from a pool of 300 candidates as our benchmark annotators.
On average, our worker selection rate stands at 18.67\%. 
Following the qualification rounds, workers are required to complete another instruction round. 
This round contains complex questions selected by the authors, and workers are required to briefly explain the answer to each question. 
The authors will then double-check the explanations provided by the annotators and disqualify those with a poor understanding.

\subsection{Annotation Instructions}
\label{appendix:annotation_interface}
For each task, we provide workers with comprehensive task explanations in layman's terms to enhance their understanding. 
We also offer detailed definitions and several examples of each choice to help annotators understand how to make decisions.
These definitions largely align with our task definitions, as explained in Section~\ref{sec:task_definitions}.
Each entry requires the worker to annotate using a four-point Likert scale. 
Workers are asked to rate each given script using such scale, where 1 signifies strong agreement and 4 indicates strong disagreement. 
We consider annotations with a value of 1 or 2 as plausible and those with a value of 3 or 4 as implausible. 

To ensure comprehension, we require annotators to confirm that they have thoroughly read the instructions by ticking a checkbox before starting the annotation task. 
We also manually monitor the performance of the annotators throughout the annotation process and provide feedback based on common errors. 
Spammers or underperforming workers will be disqualified.
The overall inter-annotator agreement (IAA) stands at 78\% in terms of pairwise agreement, and the Fleiss kappa~\cite{fleiss1971measuring} is 0.53. 
The IAA and Fleiss Kappa scores for the three subtasks are closely aligned, with a difference range of ±0.05.
These statistics are generally comparable to or slightly higher than those of other high-quality dataset construction works~\cite{DBLP:conf/aaai/SapBABLRRSC19,DBLP:conf/emnlp/FangWCHZSH21,DBLP:conf/www/FangZWSH21,DBLP:conf/aaai/HwangBBDSBC21,DBLP:journals/corr/abs-2406-02106}, which indicates that the annotators are close to achieving a strong internal agreement.

\subsection{Expert Verification}
Finally, we seek the help of three e-commerce NLP experts, each with extensive experience in NLP research, to validate the annotations. 
The experts are NLP scientists with extensive experience in e-commerce NLP. 
They are well trained in conducting NLP research and are familiar with the e-commerce domain. 
In contrast, AMT crowd-sourced workers are generally considered to have only a basic understanding of AI, NLP, and related fields. 
Therefore, recruiting experts to verify the annotated labels is critical, as they have a deeper understanding of the tasks and can better assess whether the collected labels align with the task requirements and design.
They are given the same instructions as those provided to crowd-sourcing workers and asked to verify a sample of 200 annotations for each task. 
The high level of consistency between our expert annotators and AMT annotators, as demonstrated in Table~\ref{tab:benchmark_statistics}, suggests that our AMT annotation is of high quality.

\end{document}